\documentclass[11pt]{article}

\usepackage[final]{acl}

\usepackage{times}
\usepackage{latexsym}

\usepackage[T1]{fontenc}
\usepackage[utf8]{inputenc}

\usepackage{microtype}

\usepackage{inconsolata}

\usepackage{graphicx}

\usepackage{inconsolata}

\usepackage{hyperref}
\usepackage{url}
\usepackage{booktabs}

\usepackage{lineno}
\usepackage{graphicx}

\usepackage{multirow}
\usepackage{xspace}
\usepackage{adjustbox}
\usepackage{pifont}
\usepackage{caption}
\usepackage{makecell}
\usepackage{subcaption}
\usepackage{hyperref}
\usepackage{tcolorbox}
\usepackage{lipsum} % for placeholder text
\usepackage{amsmath}
\usepackage{amssymb}
\usepackage{tablefootnote}
\usepackage{amsthm}

\usepackage[table]{xcolor}
\usepackage{array}
\usepackage[ruled,linesnumbered,vlined]{algorithm2e}

\definecolor{ourmethodbg}{RGB}{255, 250, 230}
\newcommand{\up}[1]{\textsuperscript{\textcolor{green!50!black}{$\uparrow$#1}}}
\newcommand{\down}[1]{\textsuperscript{\textcolor{red!70!black}{$\downarrow$#1}}}

\title{Mitigating Exploration Bias in RL for Multi-Instruction Following}
\author{
 \textbf{Mian Zhang\textsuperscript{1}\thanks{Equal contribution.}},
 \textbf{Yueqin Yin\textsuperscript{2}\footnotemark[1]},\\
 \textbf{Kaiyu He\textsuperscript{1}},
 \textbf{Peilin Wu\textsuperscript{1}},
 \textbf{Xinlu Zhang\textsuperscript{3}},
 \textbf{Mingyuan Zhou\textsuperscript{2}},
 \textbf{Zhiyu Zoey Chen\textsuperscript{1}}
\\
 \textsuperscript{1}UT Dallas,
 \textsuperscript{2}UT Austin,
 \textsuperscript{3}UC Santa Barbara
}

\begin{document}
\maketitle
\begin{abstract}
RL has emerged as a powerful paradigm for enhancing the instruction following capabilities of LLMs. While existing training recipes achieve substantial gains, we find that they suffer from \textbf{exploration bias towards easy instructions} when the training data has multiple instructions in a prompt. This bias is caused by two main reasons: 1) \textbf{the policy model's initial ability to satisfy hard instructions is too low} to trigger successful exploration during RL training, so the optimization is biased to easy instructions and 2) canonical RL training recipes typically employ a cumulative reward (the number of instructions fulfilled), \textbf{treating all the instructions equally}; the policy model is biased to fulfill the easy instructions to get the same amount of rewards. To address these, we first propose two metrics to measure the exploration bias of instruction following and then introduce a two-stage framework to alleviate it: 1) Behavioral Bootstrapping: a lightweight rejection sampling fine-tuning before RL to activate hard instructions and 2) Scarcity-Aware Rewards: a new RL reward function that assigns rewards to instructions based on their empirical scarcity. Experiments show that the proposed metrics are highly correlated with model performance and our methods unleash the potential of RL training: our best models outperform the baselines by a significant margin across three verifiable instruction following benchmarks. We release codes at https://github.com/mianzhang/MulIF.
\end{abstract}

% with cumulative reward

\section{Introduction}\label{sec:intro}
Since the advent of ChatGPT~\cite{Ouyang2022-ic}, instruction following has become a cornerstone capability of Large Language Models (LLMs). Reinforcement Learning (RL) has emerged as a powerful paradigm for enhancing the instruction-following capabilities of LLMs. Following \citet{Pyatkin2025-zq} and \citet{Peng2025-rv}, RL for Instruction Following (RLIF) can be formulated as optimizing a policy $\pi_{\theta}$ to generate a response $y$ that adheres to a specific set of instructions (or constraints)\footnote{We use "instructions" and "constraints" interchangeably.} $\mathcal{I} = \{I_1, I_2, \dots, I_N\}$ while fulfilling a core task (prompt) $x$. This optimization problem is modeled as maximizing the expected return of a reward function $R(y, \mathcal{I})$:$$\max_\theta \mathbb{E}_{(x, \mathcal{I}) \sim \mathcal{D}} \left[ \mathbb{E}_{y \sim \pi_\theta(\cdot|x, \mathcal{I})} [R(y, \mathcal{I})] \right]$$Here, $\mathcal{D}$ represents the distribution of prompt-instruction pairs, and $R(y, \mathcal{I})$ can be either deterministic programs or reward models. 

Canonical RLIF training recipes typically employ a cumulative reward function~\cite{Pyatkin2025-zq,Guo2025-cz}:
\begin{equation}\nonumber
R(y,\mathcal{I}) = \sum_{j=1}^{N} \mathbb{1}(I_j, y)
\end{equation}
where $\mathbb{1}(I_j, y)$ is an indicator function denoting whether the $j$-th instruction is satisfied in the rollout $y$. While intuitive, this objective introduces a \textit{exploration bias} towards easy instructions. Since easy instructions are more frequently satisfied during initial exploration, they provide a dense and stable gradient signal that dominates the optimization process. As a result, the model gradually prioritizes low-hanging fruit to maximize immediate cumulative rewards while neglecting hard instructions. This imbalance leads to a stagnation in exploring the sparse frontier of complex constraints, hindering the model's ability to achieve an  "all-pass" state in complex scenarios.

\begin{figure}[tbp]
    \centering
    \includegraphics[width=0.9\linewidth]{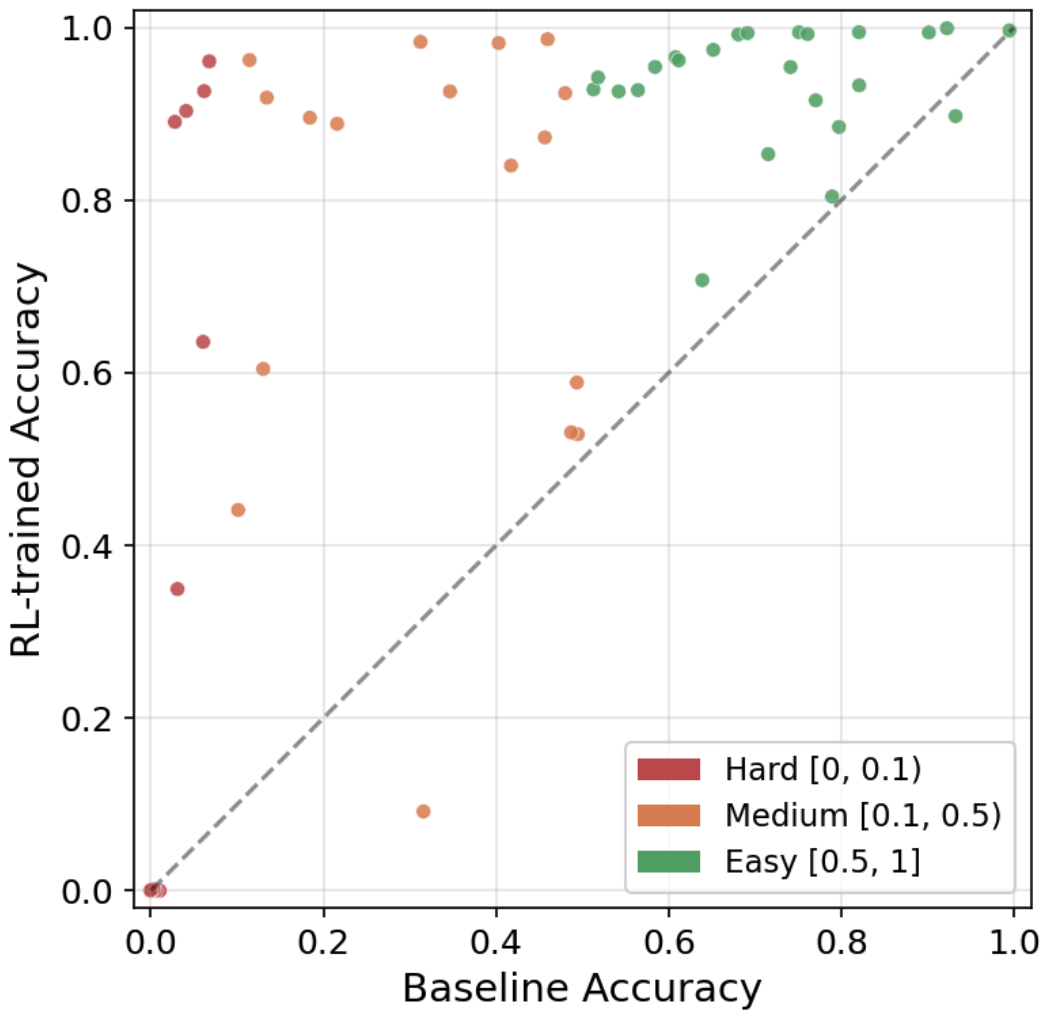}
    \caption{Instruction Accuracy: Before vs After RL}
    \label{fig:rl-vs-base-scatter}
\end{figure}

We train Qwen3-1.7B using GRPO \cite{Shao2024-dp} with the cumulative reward function on IFTrain \cite{Pyatkin2025-zq}, which consists of prompts containing up to five instructions and track the performance gain by comparing the accuracies of the base model against the RL-trained model across various instruction types. As illustrated in Figure \ref{fig:rl-vs-base-scatter}, the impact of RL training varies significantly across different instructions. We categorize instruction types into Easy, Medium, and Hard based on their baseline performance. The results provide \textit{empirical evidence} of the exploration bias: RL provides more consistent and robust improvements for Easy instructions; in contrast, performance gains for Medium and Hard instructions exhibit high variance. We also notice that \textbf{certain instructions with very low initial accuracy benefit minimally from RL training}; this suggests another reason for the exploration bias that extends beyond the limitations of the cumulative reward.

% \zhiyu{In this figure, seems there are also many medium and hard instructions with an initial low acc and trained to a high acc with RL?}

To address the exploration bias in RLIF, we first introduce two diagnostic metrics: Variation of Instruction Accuracy, which measures the imbalance in individual instruction mastery, and Variance of Synergy Accuracy, which assesses the model's uneven capacity to satisfy diverse instructions simultaneously. We experimentally demonstrate that both metrics exhibit a strong negative correlation with the prompt-level "all-pass" rate. Building on these insights, we propose a two-stage framework to alleviate the exploration bias: 1) Behavioral Bootstrapping, a lightweight Rejection Sampling Fine-Tuning phase designed to \textbf{activate hard instructions before reinforcement learning}; and 2) Scarcity-Aware Rewards, a reward mechanism that \textbf{prioritizes the discovery of rare instructions and combinations} based on their empirical scarcity. 

% which is directly inspired by our diagnostic metrics.

We conduct extensive experiments using Qwen3-1.7B and Qwen2.5-7B-Instruct across three verifiable instruction following benchmarks. Our results demonstrate that models integrated with Behavioral Bootstrapping and Scarcity-Aware Rewards significantly outperform the standard cumulative reward baseline by an average of 4.5 points across all evaluated benchmarks and achieve substantial gains of up to 9.2 points on our primary test set, IFBench, enabling these relatively small models to match the performance of much larger off-the-shelf models such as Llama-3.1-70B and GPT-OSS-20B. Finally, we conduct ablation studies to reveal the impact of our proposed framework under varying configurations, providing practical insights for their optimal implementation.

% \ref{fig:rl-vs-base-scatter}.

\section{Related Work}

\subsection{Instruction Following}

Precisely following instructions is a fundamental capability of LLMs \cite{Zhou2023-bp, Pyatkin2025-zq}. Unlike general instruction tuning \cite{Chung2022-ls, Wang2022-ux}, which focuses on teaching models basic skills to understand user intent and solve tasks, precise instruction following is a more rigorous challenge. It requires LLM agents to adhere strictly to specific constraints within a prompt, a factor critical for the reliability and evolution of autonomous agents \cite{Wu2024-le,Fu2025-la}. As tasks grow in complexity, the number of instructions per task continues to increase \cite{Qi2025-sn, He2025-zu, Pyatkin2025-zq}, yet empirical evidence suggests that model performance often degrades as constraint density rises \cite{Jaroslawicz2025-pb, Guo2025-cz}. RL has proven effective for improving instruction following \cite{Bercovich2025-tt,Pyatkin2025-zq,Peng2025-rv}. 
Our work focuses on alleviating the exploration bias in RL optimization when the model is confronted with multiple, diverse instructions.

\subsection{Bias in RL Training}
RL training of LLMs is known to introduce systematic biases. A well-studied instance is \textit{length bias}, where policies exploit verbosity as a reward shortcut~\cite{singhal2024long, chen2024odin}. RL fine-tuning also reduces output diversity through \textit{mode collapse}~\cite{kirk2024understanding}, and in the verifiable-reward regime, \textit{entropy collapse} causes the policy to lose exploratory capacity early in training~\cite{yu2026dapo}. When training involves multiple objectives or tasks, optimization imbalance becomes more pronounced. \citet{wu2026imbalanced} show that gradient magnitudes across tasks can differ by up to $33\times$ in RL post-training, causing easy tasks to dominate, and multi-objective RLHF methods must carefully balance conflicting alignment objectives to avoid one dominating others~\citep{dai2024safe, moskovitz2024confronting}. Recent work on RLVR has further identified \textit{exploration bottlenecks}, where low initial accuracy on hard problems yields sparse rewards and creates a self-reinforcing loop that prevents learning~\cite{chen2026does}. While these works address biases at the reward, algorithm, or task level, the exploration bias \textit{within} a single prompt, where instructions of varying difficulty coexist, remains unexplored. 

% Our work identifies this intra-prompt bias, proposes metrics to quantify it, and introduces targeted methods to mitigate it.
% and RL systematically prioritizes easy ones,
\section{Measure of Exploration Bias}\label{sec:measure}

Given a prompt with a set of instructions $\mathcal{I} = \{I_1, I_2, \dots, I_N\}$, we define the marginal satisfaction probability $P(I_j)$ as the probability that a response $y$ sampled from the policy $\pi_{\theta}$ satisfies $I_j$:
\begin{equation}\nonumber
P(I_j) = \mathbb{P}_{y \sim \pi_{\theta}} (\mathbb{1}(I_j, y) = 1)
\end{equation}
Standard evaluations of instruction following typically focus on maximizing \textit{prompt accuracy} or more granular \textit{instruction accuracy}. However, neither metric adequately captures the model's exploration bias. We propose two metrics based on the intuition that a model should maintain balanced capability in satisfying instructions and synergizing possible instruction pairs within a single response.

\paragraph{Variation of Instruction Accuracy (VIA).} 
VIA serves as a metric for the imbalance in a model's individual instruction mastery. Given a prompt with $N$ instructions, VIA is calculated as the \textit{variance of the marginal satisfaction probabilities $\{P(I_j)\}_{j=1}^N$ across the instruction set}. A high VIA indicates that the model consistently excels at certain instructions while failing others.

\paragraph{Variance of Synergy Accuracy (VSA).} 
While VIA focuses on individual instructions, VSA measures the imbalance in the model's capability to synergize instruction pairs. We first define the pairwise synergy probability $P(I_j, I_k)$ for any pair of instructions $(I_j, I_k)$ as the ratio of their joint satisfaction probability to their individual bottleneck:
\begin{equation}\label{eq:pair}
P(I_j, I_k) = \frac{P(I_j \cap I_k)}{\min(P(I_j), P(I_k))}
\end{equation}
where $P(I_j \cap I_k)$ is the joint probability that a single response satisfies both $I_j$ and $I_k$. 
VSA is then calculated as the \textit{variance of the pairwise synergy probabilities across all unique instruction pairs $\mathcal{P} = \{(j, k) \mid 1 \leq j < k \leq N\}$ where $P(I_j) > 0$ and $P(I_k) > 0$}. A high VSA indicates that the model is significantly better at synergizing certain instruction pairs while failing others.

\begin{figure}[tbp]
    \centering
    \includegraphics[width=0.85\linewidth]{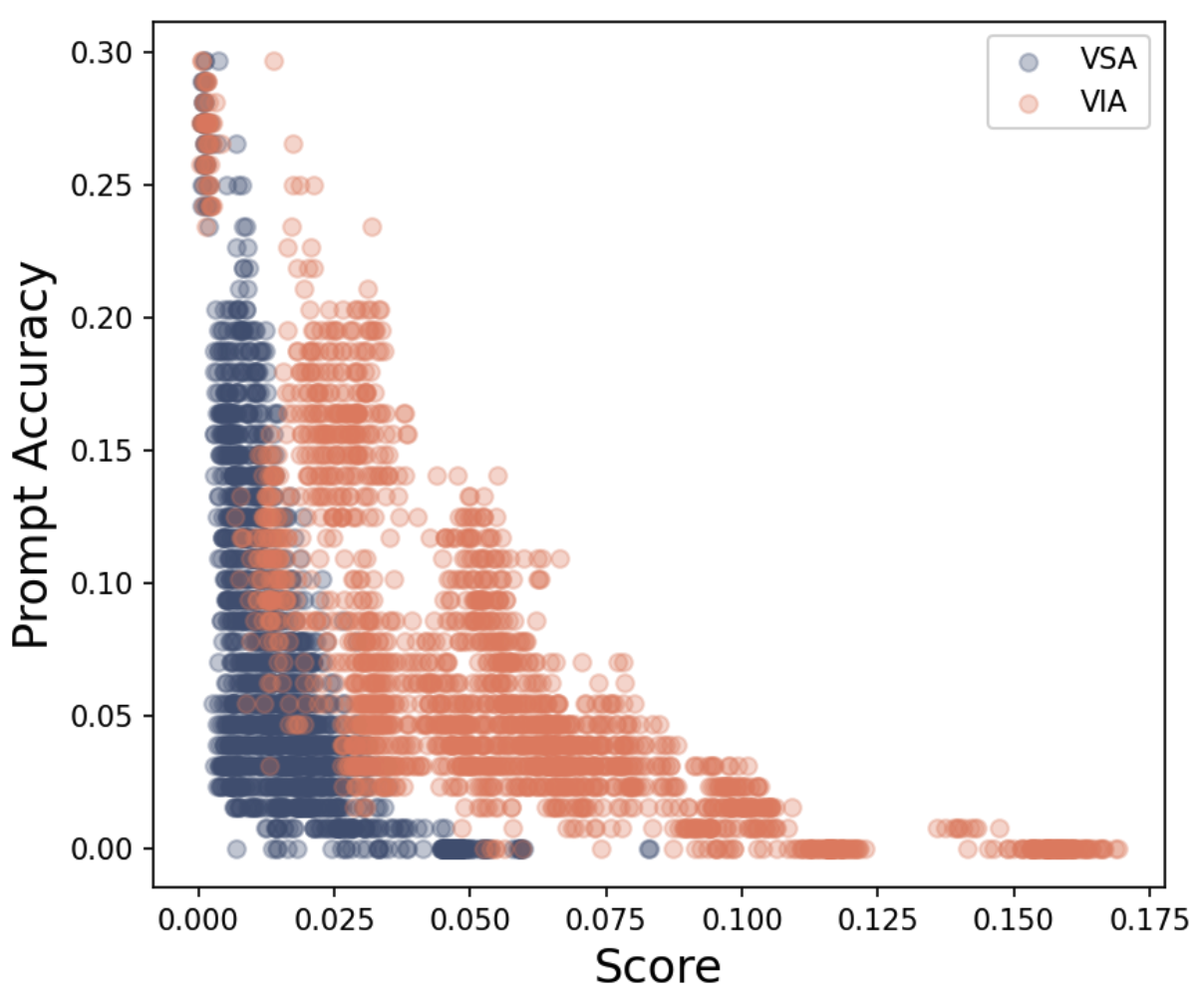}
    \caption{VIA and VSA vs Prompt Accuracy}
    \label{fig:via_vsa}
\end{figure}
\begin{figure*}[tbp]
    \centering
    \includegraphics[width=\textwidth]{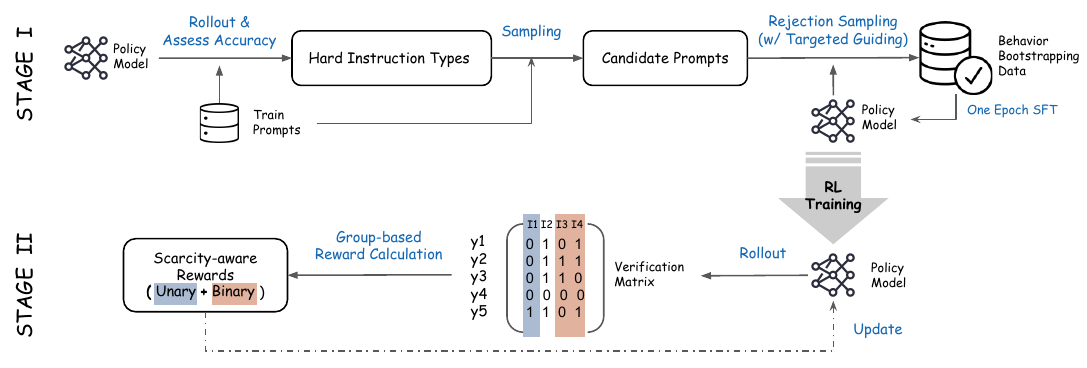}
    \caption{Overview of the framework for mitigating exploration bias. \textbf{Top} (Behavioral Bootstrapping): The policy model is first evaluated on the training set to identify hard instruction types. Prompts containing at least one hard instruction are then sampled, and a subsequent rejection sampling phase is conducted to collect bootstrapping data. Finally, the policy model is supervised fine-tuned for one epoch on this collected data. \textbf{Bottom} (Scarcity-Aware Rewards): The bootstrapped model is fine-tuned via RL using Scarcity-Aware Rewards (Unary + Binary), which are calculated across a response group to incentivize the exploration of rare instructions and combinations.}
    \label{fig:method}
\end{figure*}

\paragraph{Relationship with Prompt Accuracy} To investigate the relationship between VIA, VSA, and prompt accuracy, we conduct a controlled experiment. We utilize test prompts from AdvancedIF~\cite{He2025-zu}, which contain diverse multiple instructions for each prompt. We construct a comprehensive response pool by aggregating outputs from a wide range of popular models, including GPT-5.4, GPT-5.4-mini, GPT-5, GPT-4o, GPT-OSS-120B, GPT-OSS-20B, Qwen-3.5-35B-A3B, Qwen-3.5-9B, Qwen-3-32B, Qwen-3-8B, Qwen-3-1.7B, Olmo-3.1-32B-Instruct, Olmo-3-7B-Instruct, Llama-3.2-3B, and Llama-3.1-8B. We then repeatedly sample response groups from this pool and \textbf{only keep those satisfying a fixed number of instructions} (same instruction accuracy). For each sampled group of size $K$, we derive a verification matrix $\mathbf{M} \in \{0, 1\}^{K \times N}$, where $\mathbf{M}_{i,j} = 1$ if the $i$-th response satisfies the $j$-th instruction, and $0$ otherwise. Based on $\mathbf{M}$, we estimate the marginal satisfaction probability $P(I_j)$ and the pairwise synergy probability $P(I_j \cap I_k)$ using the following empirical frequencies:
\begin{equation}\nonumber
P(I_j) \approx \frac{n_j}{K}, \quad P(I_j \cap I_k) \approx \frac{m_{j,k}}{K}
\end{equation}
where $n_j = \sum_{i=1}^K \mathbf{M}_{i,j}$ represents the success count for instruction $I_j$, and $m_{j,k} = \sum_{i=1}^K (\mathbf{M}_{i,j} \cdot \mathbf{M}_{i,k})$ denotes the joint success count for the instruction pair $(I_j, I_k)$. The pairwise synergy probability is calculated by substituting these frequency-based estimates into Eq.~\ref{eq:pair}. We employ a group size of $K=128$ and fix the instruction accuracy at $0.68 \pm 0.01$, which corresponds to the mean instruction accuracy across the evaluated models. As illustrated in Figure~\ref{fig:via_vsa}, \textbf{both VIA and VSA exhibit a consistent inverse relationship with prompt accuracy}: a reduction in variance, whether in individual instruction mastery or pairwise synergy, strongly correlates with higher prompt-level success rates. These results empirically validate that \textbf{models with minimized exploration bias are significantly more likely to fulfill the entire set of instructions within a prompt simultaneously}.

% Consequently, VIA and VSA serve as robust diagnostic metrics for monitoring and quantifying exploration bias in multi-instruction scenarios.

% We set the group size as 128 and a fixed instruction acc of 0.68 with a duration of 0.01, which is the average instruction accuracy of the tested models. Figure~\ref{fig:via} and \ref{fig:vsa} show the relationship of VIA, VSA between prompt accuracy, we could see that both VIA and VSA share similar relationship with prompt accuracy that lower variance leads to higher prompt accuracy. This indicates \textbf{models with less exploration bias are more likely to fulfill all the instructions in a prompt}. And VIA and VSA could well monitor the exploration bias.  

\section{Mitigating Exploration Bias}
As mentioned in Section \ref{sec:intro}, two factors contribute to the exploration bias of RLIF: the policy model's low initial capability regarding certain instructions and a standard cumulative reward formulation that treats all instructions equally. To alleviate these, we propose a two-stage framework including 1) lightweight behavioral bootstrapping to first activate difficult instructions and 2) a new reward design inspired by the metrics proposed in Section \ref{sec:measure} that assigns rewards based on the empirical scarcity of instructions. Figure \ref{fig:method} provides an overview.
% \zhiyu{Figure 3 has too much space. Can make it more compact.}
 % in RL for multi-instruction following

\subsection{Behavioral Bootstrapping}
In multi-instruction scenarios, complex constraints often suffer from low initial accuracy, rendering standard RL ineffective due to the lack of successful exploration. To bridge this gap, we introduce Behavioral Bootstrapping via a lightweight Rejection Sampling Fine-Tuning phase. We identify seed instruction types $\mathcal{T}_{seed}$ that fall below a predefined accuracy threshold $\tau$ and curate a minimal dataset $\mathcal{D}_{seed}$ of successful rollouts for them. For each target type $t$, we sample $K$ responses per prompt, selecting those that satisfy the target instruction while maximizing the coverage of other instructions in a prompt. If the base policy fails to generate a valid response, we employ a \textit{Targeted Guiding} prompting by appending a suffix \textit{"Focus on fulfilling [Instruction] first"} to nudge the model toward the target instruction. After collecting $\mathcal{D}_{seed}$ with at most $N_{target}$ for each selected instruction type, the initial policy is supervised fine-tuned on $\mathcal{D}_{seed}$ for a single epoch to boost the initial capability of hard instructions. This establishes an ideal base for subsequent RL training. The whole process is detailed in Algo. \ref{alg:bootstrapping}.
% 定义专业调色盘
\definecolor{algoKeyword}{RGB}{0,0,200}    
\definecolor{algoComment}{RGB}{34,139,34} 
\definecolor{algoSteps}{RGB}{139,0,0}   
\definecolor{algoInput}{RGB}{80,80,80} 

% 配置 algorithm2e 的颜色
\renewcommand{\KwSty}[1]{\textcolor{algoKeyword}{\textbf{#1}}}
\renewcommand{\ArgSty}[1]{\textnormal{#1}}
\SetKwComment{Comment}{\textcolor{algoComment}{$\triangleright$\ }}{}

\begin{algorithm}[t!]
\small
\caption{Behavioral Bootstrapping}
\label{alg:bootstrapping}

\SetKwInOut{Input}{\textcolor{algoInput}{Input}}
\SetKwInOut{Output}{\textcolor{algoInput}{Output}}

\SetInd{0.4em}{1em}

\Input{Initial policy $\pi_{\theta}$, Data $\mathcal{D}_{train}$, Threshold $\tau$, Sample size $K$, Target count $N_{target}$}
\Output{Bootstrapped policy $\pi_{\theta'}$}
\BlankLine

{\small\itshape\textcolor{algoComment}{$\triangleright$ Step 1: Evaluation of initial policy}}
% \textcolor{algoSteps}{\textbf{Step 1:}} Evaluation of initial policy

Evaluate $\pi_{\theta}$ on $\mathcal{D}_{train}$ to compute accuracy $\text{Acc}(t)$ for each instruction type $t \in \mathcal{T}$ \;

Identify seed types $\mathcal{T}_{seed} = \{ t \in \mathcal{T} \mid \text{Acc}(t) < \tau \}$\;
$\mathcal{D}_{seed} \gets \emptyset$\;

\BlankLine

{\small\itshape\textcolor{algoComment}{$\triangleright$ Step 2: Seed data curation via RSFT}}
% \textcolor{algoSteps}{\textbf{Step 2:}} Seed data curation via RSFT

\ForEach{instruction type $t \in \mathcal{T}_{seed}$}{
    \While{$|\mathcal{D}_{seed, t}| < N_{target}$}{
        
        Sample a prompt $x$ from $\mathcal{D}_{train}$ containing an instruction $I \in t$ \;
        Generate $\mathcal{Y} = \{y_1, \dots, y_K\} \sim \pi_{\theta}(\cdot|x)$\;
        $\mathcal{Y}_{valid} = \{ y \in \mathcal{Y} \mid y \text{ satisfies } I \}$\;
        
        \BlankLine
        
        \If{$\mathcal{Y}_{valid} = \emptyset$}{
            {\small\itshape\textcolor{algoComment}{$\triangleright$ Targeted guiding}}
            
            $x\gets x+\textit{“Focus on fulfilling [I] first.”}$ \;
            
            $\mathcal{Y}_{valid} \gets \{ y \sim \pi_{\theta}(\cdot|x) \mid y \text{ satisfies } I \}$ \;
        }
        
        \BlankLine
        
        \If{$\mathcal{Y}_{valid} \neq \emptyset$}{
            {\small\itshape\textcolor{algoComment}{$\triangleright$ Maximize auxiliary coverage}}
            % $y^* = \arg\max_{y \in \mathcal{Y}_{valid}} | \{ I' \in \text{Prompt}(x) \mid y \text{ satisfies } I' \} |$ \;
            
            Select $y^*$ with highest instruction coverage\;
            $\mathcal{D}_{seed} \gets \mathcal{D}_{seed} \cup \{(x, y^*)\}$ \;
        }
    }
}

\BlankLine

Supervised fine-tune $\pi_{\theta}$ on $\mathcal{D}_{seed}$ for 1 epoch $\rightarrow \pi_{\theta'}$\;
\end{algorithm}
% \vspace{-5em}
\definecolor{headerblue}{RGB}{189, 215, 238} % 主表头蓝色
\definecolor{rowblue}{RGB}{221, 235, 247}    % 分类行浅蓝色
\definecolor{ourmethodbg}{RGB}{255, 250, 230} % 突出显示“我们的方法”
\definecolor{posgreen}{RGB}{0, 150, 0}        % 提升用绿色
\definecolor{negred}{RGB}{200, 0, 0}          % 降低用红色

% Add these definitions to your document preamble if not already present:
% \usepackage[table]{xcolor}
% \newcommand{\up}[1]{\textsuperscript{\textcolor{green!50!black}{$\uparrow$#1}}}
% \newcommand{\down}[1]{\textsuperscript{\textcolor{red!70!black}{$\downarrow$#1}}}

\begin{table*}[h]
    \centering
    \small
    \renewcommand{\arraystretch}{1.2}
    \setlength{\tabcolsep}{5pt}
    \begin{tabular}{l cc cc ccc c}
        \toprule
        & \multicolumn{2}{c}{\textit{\textbf{IFBench}}} & \multicolumn{2}{c}{\textit{\textbf{IFEval}}} & \multicolumn{3}{c}{\textit{\textbf{Multi-IF}}} & \\
        \cmidrule(lr){2-3} \cmidrule(lr){4-5} \cmidrule(lr){6-8}
        \textbf{Model} & \textbf{Loose} & \textbf{Strict} & \textbf{Loose} & \textbf{Strict} & \textbf{Turn 1} & \textbf{Turn 2} & \textbf{Turn 3} & \textbf{Average} \\
        \midrule
        \rowcolor{rowblue} \multicolumn{9}{l}{\textbf{Off-the-shelf}} \\
        GPT-OSS-120B & 65.2 & 59.6 & 86.6 & 83.7 & 76.4 & 65.1 & 54.8 & 70.2 \\
        Llama-3.1-70B & 35.0 & 29.8 & 83.7 & 81.8 & 80.5 & 71.9 & 63.1 & 63.7 \\
        Qwen3-32B & 31.9 & 28.0 & 89.0 & 86.1 & 84.1 & 74.3 & 60.6 & 64.8 \\
        Gemma-3-27B & 33.3 & 25.6 & 82.9 & 80.2 & 82.5 & 72.9 & 63.7 & 63.0 \\
        GPT-OSS-20B & 40.0 & 37.5 & 82.8 & 79.6 & 61.4 & 52.4 & 42.1 & 56.5 \\
        Qwen3-8B & 31.2 & 26.6 & 85.7 & 82.4 & 82.6 & 73.1 & 64.4 & 63.7 \\
        Mistral-7B-Instruct-v0.3 & 17.1 & 14.3 & 45.8 & 42.3 & 44.0 & 30.5 & 19.6 & 30.5 \\
        Gemma-3-4B & 28.4 & 22.4 & 77.0 & 73.1 & 76.3 & 63.7 & 53.0 & 56.3 \\
        \midrule
        \rowcolor{rowblue} \multicolumn{9}{l}{\textbf{Qwen3-1.7B (Qwen1.7B)}} \\
        Qwen1.7B & 23.1 & 18.5 & 70.2 & 67.6 & 67.3 & 55.2 & 44.2 & 49.4 \\
        Qwen1.7B-CR & 41.1 & 35.1 & 86.8 & 84.1 & 81.2 & 59.2 & 46.7 & 62.0 \\
        \rowcolor{ourmethodbg} Qwen1.7B-BeBoot-CR$^*$ & 43.1\up{2.0} & 39.2\up{4.1} & 90.0\up{3.2} & 88.3\up{4.2} & 78.3\down{2.9} & 61.5\up{2.3} & 48.9\up{2.2} & 64.2\up{2.2} \\
        \rowcolor{ourmethodbg} Qwen1.7B-BeBoot-SaR$^*$ & \textbf{50.1}\up{9.0} & \textbf{44.3}\up{9.2}  & \textbf{90.7}\up{3.9} & \textbf{89.0}\up{4.9} & \textbf{85.4}\up{4.2} & \textbf{62.9}\up{3.7} & \textbf{50.4}\up{3.7} & \textbf{67.5}\up{5.5} \\
        \midrule
        \rowcolor{rowblue} \multicolumn{9}{l}{\textbf{Qwen2.5-7B-Instruct (Qwen7B)}} \\
        Qwen7B & 30.5 & 27.7 & 75.0 & 71.9 & 75.0 & 59.0 & 47.6 & 55.2 \\
        Qwen7B-CR & 42.4 & 39.5 & 91.1 & 90.0 & \textbf{88.0} & 68.2 & 55.9 & 67.9 \\
        \rowcolor{ourmethodbg} Qwen7B-BeBoot-CR$^*$ & 48.4\up{6.0} & 43.1\up{3.6} & 92.6\up{1.5} & 91.2\up{1.2} & 84.7\down{3.3} & 68.3\up{0.1} & \textbf{57.7}\up{1.8} & 69.4\up{1.5} \\
        \rowcolor{ourmethodbg} Qwen7B-BeBoot-SaR$^*$ & \textbf{52.6}\up{10.2} & \textbf{48.4}\up{8.9} & \textbf{94.0}\up{2.9} & \textbf{92.9}\up{2.9} & 86.5\down{1.5} & \textbf{69.3}\up{1.1} & 56.6\up{0.7} & \textbf{71.5}\up{3.6} \\
        \bottomrule
        % \multicolumn{9}{l}{\footnotesize $^*$ Denotes our method. Baseline: Qwen1.7B-CR, Qwen7B-CR}
    \end{tabular}
    \caption{Performance on Instruction Following Benchmarks. All off-the-shelf model results are reproduced by us. For the Qwen3 series, we use their non-thinking mode, while for GPT-OSS models, we use the default (medium) thinking effort. The best performance within each column for the same base model is boldfaced and the performance comparison of \colorbox{ourmethodbg}{our methods} with the baselines (Qwen1.7B-CR and Qwen7B-CR) is annotated.}\label{tab:main}
\end{table*}
\subsection{Scarcity-Aware Rewards}
Standard cumulative reward treats instructions equally during RL training. However, instructions vary in difficulty, causing the model to develop a bias toward exploring easier instructions because fulfilling an easy instruction receives the same reward as a difficult one. To address this, inspired by the metrics introduced in Section \ref{sec:measure}, we design a new reward function that encourages the model to explore rare instructions or rare combinations. 

The intuition of our reward design is \textbf{hard instructions satisfaction should receive more rewards and easy ones should receive less} and we use \textit{empirical scarcity} in a response group during training to estimate the difficulty. Specifically, we first let the model generate $K$ responses for a prompt containing a set of instructions $\mathcal{I} = \{I_1, I_2, \dots, I_N\}$, and employ verification functions to obtain a verification matrix $\mathbf{M} \in \{0, 1\}^{K \times N}$. We then estimate the model’s marginal satisfaction probability for an instruction, $P(I_j)$, and the pairwise synergy probability, $P(I_j, I_k)$, based on current model performance (see Section \ref{sec:measure}). The reward for instruction $I_j$ is defined as $1 - P(I_j)$, and for an instruction pair as $1 - P(I_j, I_k)$. This design ensures that RL optimization treats instructions differently. The final reward for rollout $R_i$ is formulated as:
\begin{equation}\label{eq:reward}
\begin{aligned}
\mathcal{R}(R_i) &= \underbrace{\frac{1}{N}\sum_{j=1}^N \mathbf{M}_{i,j} \left( 1 - \frac{n_j}{K}\right)}_{\text{Unary Reward}} \\
&+ \underbrace{\frac{\alpha}{|\mathcal{P}_i|} \sum_{(j,k) \in \mathcal{P}_i} \left( 1 - \frac{\sum_{m=1}^K \mathbf{M}_{m,j} \mathbf{M}_{m,k}}{\min(n_j, n_k)} \right)}_{\text{Binary Reward}}
\end{aligned}
\end{equation}
where $n_j = \sum_{i=1}^K \mathbf{M}_{i,j}$ denote the total successes for instructions $j$ in the response group. $\mathcal{P}_i = \{(j,k) \mid j < k, \mathbf{M}_{i,j} \cdot \mathbf{M}_{i,k} = 1\}$ is the set of satisfied instruction pairs in a response. If $n_j=0$ or $n_k=0$, the binary reward is set to 0. $\alpha$ is a balancing factor. The Unary Reward scales the contribution of each satisfied instruction by its empirical scarcity; instructions that are rarely satisfied by the current policy receive higher weights. The Binary Reward focuses on the difficulty of instruction "co-occurrence." By normalizing the intersection against the minimum of individual successes $\min({n_j},n_k)$, it specifically penalizes pairs that are easy to satisfy separately but difficult to satisfy together, thereby encouraging the model to overcome synergy bottlenecks. Please note that while our proposed reward can theoretically be integrated into any policy gradient method, group-based approaches like GRPO \cite{Shao2024-dp} are a more natural choice because they utilize a response group to estimate advantages for optimization. Our reward can be derived directly from these samples with negligible additional cost.

% The new reward is naturally can be integrated group-based policy gradient method like GRPO. 

% based on scarcity, encouraging exploration of rarely satisfied combinations.
\section{Experiments}
\subsection{Experimental Settings}
We select \textit{verifiable instruction following} as the testbed to evaluate our proposed methods. This choice is motivated by the fact that verification can be executed via deterministic functions, yielding 100\% accuracy. In contrast, soft constraints require a model-based verifier, which frequently suffers from reliability issues, even when state-of-the-art models are employed \cite{Wen2026-wz}, and thus interfering with the evaluation process.
\paragraph{Training} We utilize the Qwen3-1.7B \cite{Yang2025-bp} and Qwen2.5-7B-Instruct \cite{Qwen2024-mx} as the primary base models for our experiments. Training is conducted using IFTrain \cite{Pyatkin2025-zq}, a dataset with prompts containing rule-based instructions. To specifically investigate model behavior in multi-instruction scenarios, we filter the original dataset to retain only prompts containing 3 to 5 instructions, resulting in a specialized training set of 45,374 prompts. Regarding Behavioral Bootstrapping, we use an instruction selection threshold $\tau$ of 0.5 and the target seed count $N_{target}$ of 100 ($\mathcal{D}_{seed}=900$). For the RL phase, we employ GRPO with a response group size of 16 and a total batch size of 512. We use a learning rate of $1 \times 10^{-6}$ and set the reward balance factor $\alpha$ (Eq. \ref{eq:reward}) to 2. Based on initial runs, the maximum training duration is set to 400 steps. All experiments are executed on four NVIDIA A100 GPUs. For further details on the sensitivity of $\tau$, $N_{target}$ and $\alpha$, please refer to Section~\ref{sec:ans}.

% the ablation studies in
\paragraph{Evaluation} The model's performance is evaluated on three precise instruction following benchmarks: IFBench~\cite{Pyatkin2025-zq}, IFEval~\cite{Zhou2023-bp}, and Multi-IF \cite{He2024-vi}. IFBench serves as our primary test set, featuring out-of-domain instruction types relative to the training data. In contrast, IFEval functions as an in-domain benchmark, as its instruction types overlap with those in the training set. Multi-IF assesses the models' proficiency in following multi-turn and multilingual instructions. Consistent with the settings recommended by the Qwen team for optimal output quality, all generations are conducted with a temperature of 0.7. We follow the original papers to report loose and strict prompt accuracy for IFBench and IFEval and an average of four types of accuracy scores on each turn.

% , top\_p = 0.8, and top\_k = 20
% InFoBench and AdvancedIF target "soft" instructions, which necessitate the use of a model-based checker for verification; specifically, we employ GPT-OSS-120B as the checker in this work.
% InFoBench~\cite{Qin2024-cs} and AdvancedIF~\cite{He2025-zu}

\begin{figure*}[htbp]
    \centering
    \includegraphics[width=\linewidth]{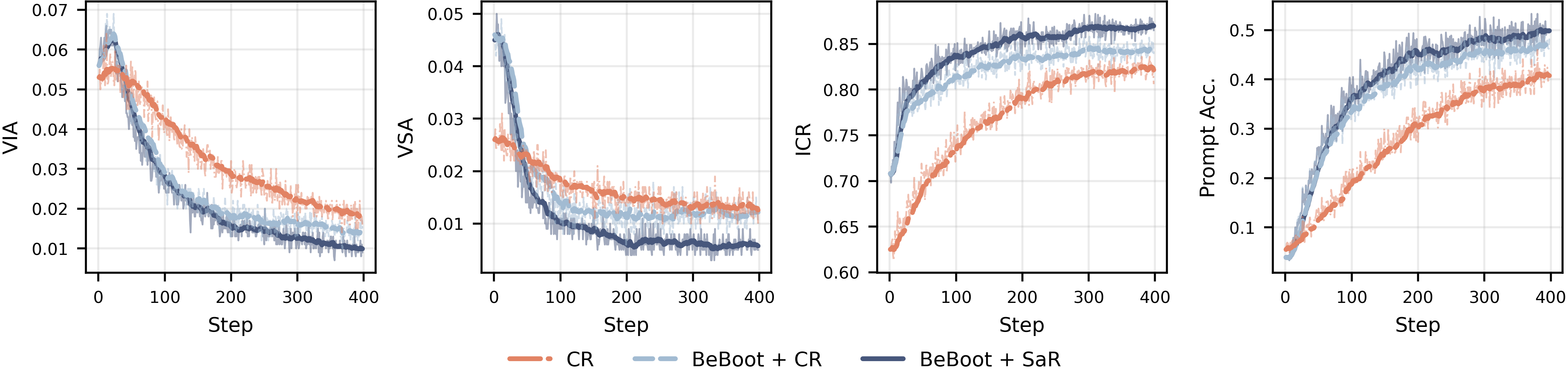}
    \vspace{-1.5em}
    \caption{The training curves of VIA, VSA, ICR, and Prompt Accuracy.}
    \label{fig:via_vsa_icr}
\end{figure*}
\begin{figure}[h!]
    \centering
    \includegraphics[width=0.8\linewidth]{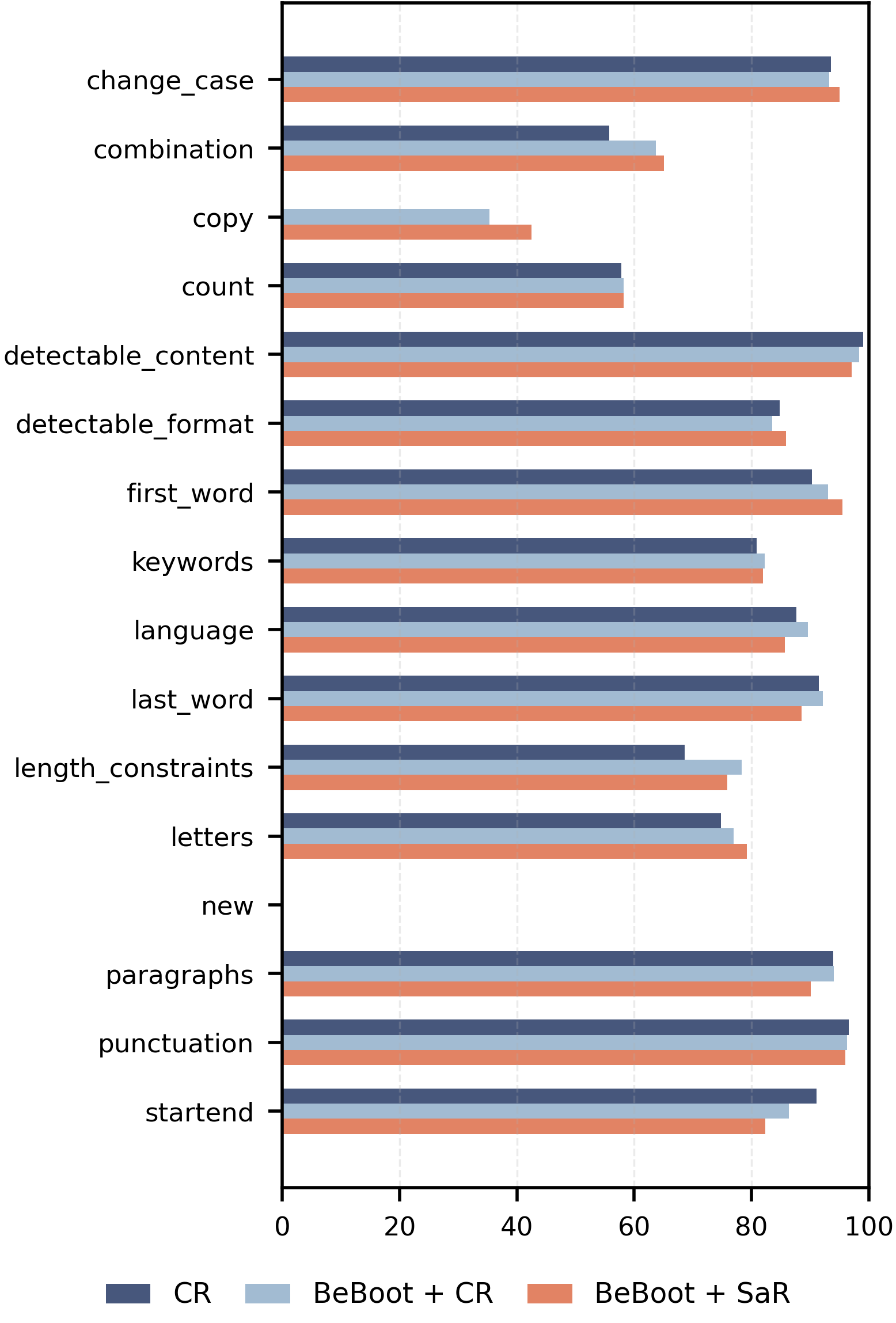}
    \caption{Accuracy Shift by Instruction Type}
    \label{fig:dist}
\end{figure}
\subsection{Main Results}
% For better illustration, we denote Culmulative Reward as CR, the proposed Behaviour BootStrapping as BeBoot and Scarcity-aware Rewards as SaR.
% \zhiyu{In Table 1, which row is ours? Mark it out as `ours'.}
\paragraph{Model Performance} Table~\ref{tab:main} shows the model performance on the benchmarks. The proposed mitigation methods for exploration bias, Behavioral Bootstrapping (BeBoot) and Scarcity-Aware Rewards (SaR), significantly enhance the instruction-following capabilities of the RL-trained models, outperforming the cumulative reward (CR) baselines \textbf{with an average gain of 5.5 on 1.7B base model and 3.6 on 7B base model across the evaluated benchmarks}. Most notably, our best models, Qwen1.7B-BeBoot-SaR and Qwen7B-BeBoot-SaR, achieve scores of 44.3 and 48.4 strict accuracy, respectively, \textbf{with substantial gains of 9.2 and 8.9 points over the baselines on our primary test set IFBench}. This indicates that the models trained with BeBoot and SaR have better generalization to unknown instruction types. This allows our small models to match the overall performance of much larger off-the-shelf models like Llama-3.1-70B and GPT-OSS-120B. We can also see that while BeBoot provides a critical initial boost by addressing early exploration failures, the integration of SaR proves essential for peak performance, as evidenced by the additional performance increase when comparing -BeBoot-SaR to -BeBoot-CR. These findings validate that mitigating exploration bias through targeted bootstrapping and scarcity-aware RL is effective for mastering complex instruction sets. 

% There is drop on AdvancedIF for Qwen1.7B when applying BeBoot and SaR. We attributes it to the distribution discrepancy between the training data and AdvancedIF.

\paragraph{Exploration Bias} The training curves in Figure \ref{fig:via_vsa_icr} illustrate how our proposed framework mitigates exploration bias compared to the standard cumulative reward (CR) baseline. We estimate the Variation of Instruction Accuracy (VIA) and Variance of Synergy Accuracy (VSA) for each prompt based on the response groups as described in Section \ref{sec:measure}. Additionally, we visualize the Instruction Coverage Rate (ICR), which measures the proportion of instructions fulfilled at least once by the response group, and prompt accuracy. We can see that while the standard CR approach exhibits a slow, gradual reduction in VIA and VSA, it consistently maintains higher levels of instructional imbalance. In contrast, methods utilizing BeBoot begin at a significantly higher ICR and prompt accuracy, successfully activating hard instructions that standard reinforcement learning typically neglects. Notably, \textbf{BeBoot+SaR achieves the fastest and most substantial decline in both VIA and VSA, reaching the lowest variance levels across all tested methods}. This reduction signifies a more balanced exploration of the instruction space, which directly facilitates the model's ability to satisfy multiple diverse instructions simultaneously. The trajectories for ICR and prompt accuracy further confirm that \textbf{BeBoot+SaR enables the model to address more instructions per prompt, resulting in superior prompt-level success on the training set}. Figure~\ref{fig:dist} provides a detailed decomposition of the pass rate shift across instruction types. We observe that \textbf{BeBoot and SaR yield significant improvements over the CR baseline on challenging categories} such as \texttt{combination}, \texttt{copy}, and \texttt{length\_constraint}. Furthermore, the integration of both BeBoot and SaR outperforms the individual application of BeBoot. We also notice that all models achieve a $0\%$ pass rate on the \texttt{new} category; this is likely due to extremely difficult instructions, such as: \textit{"Copy the span of words that lies between (and including) index 13 and 43, where the indices refer to character positions."}

% These trends confirm that minimizing exploration bias through scarcity-aware and synergy-focused rewards is critical for reaching peak prompt-level performance in complex scenarios.
\begin{figure}[h!]
    \centering
    \includegraphics[width=0.9\linewidth]{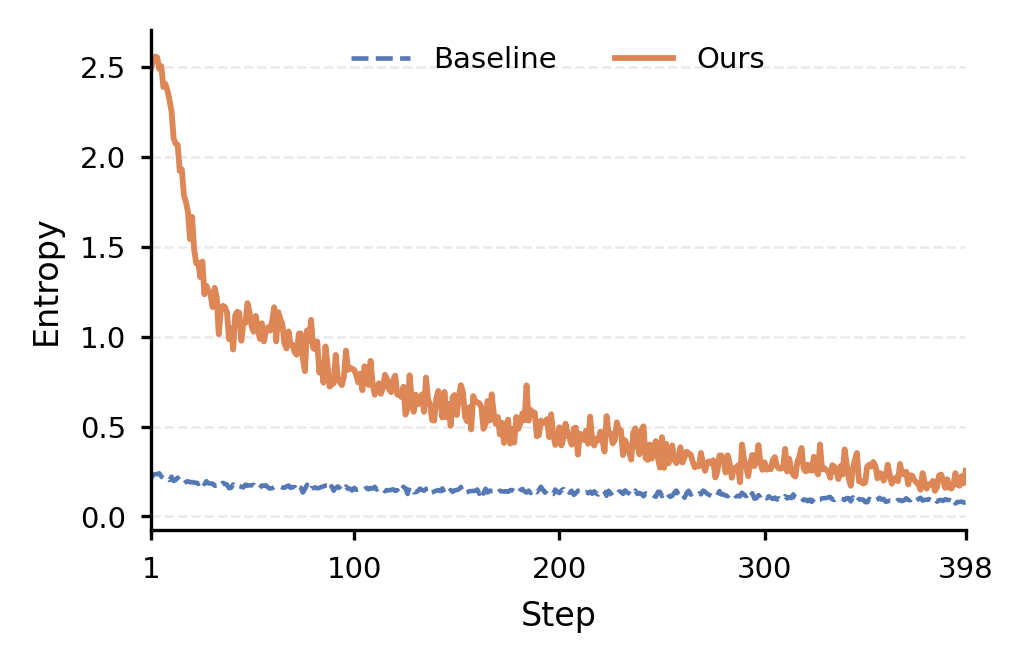}
    \vspace{-1em}
    \caption{Policy Entropy}
    \label{fig:entropy}
\end{figure}
\begin{figure*}[t]
  \centering
  \begin{subfigure}[t]{0.33\textwidth}
    \centering
    \includegraphics[width=\linewidth]{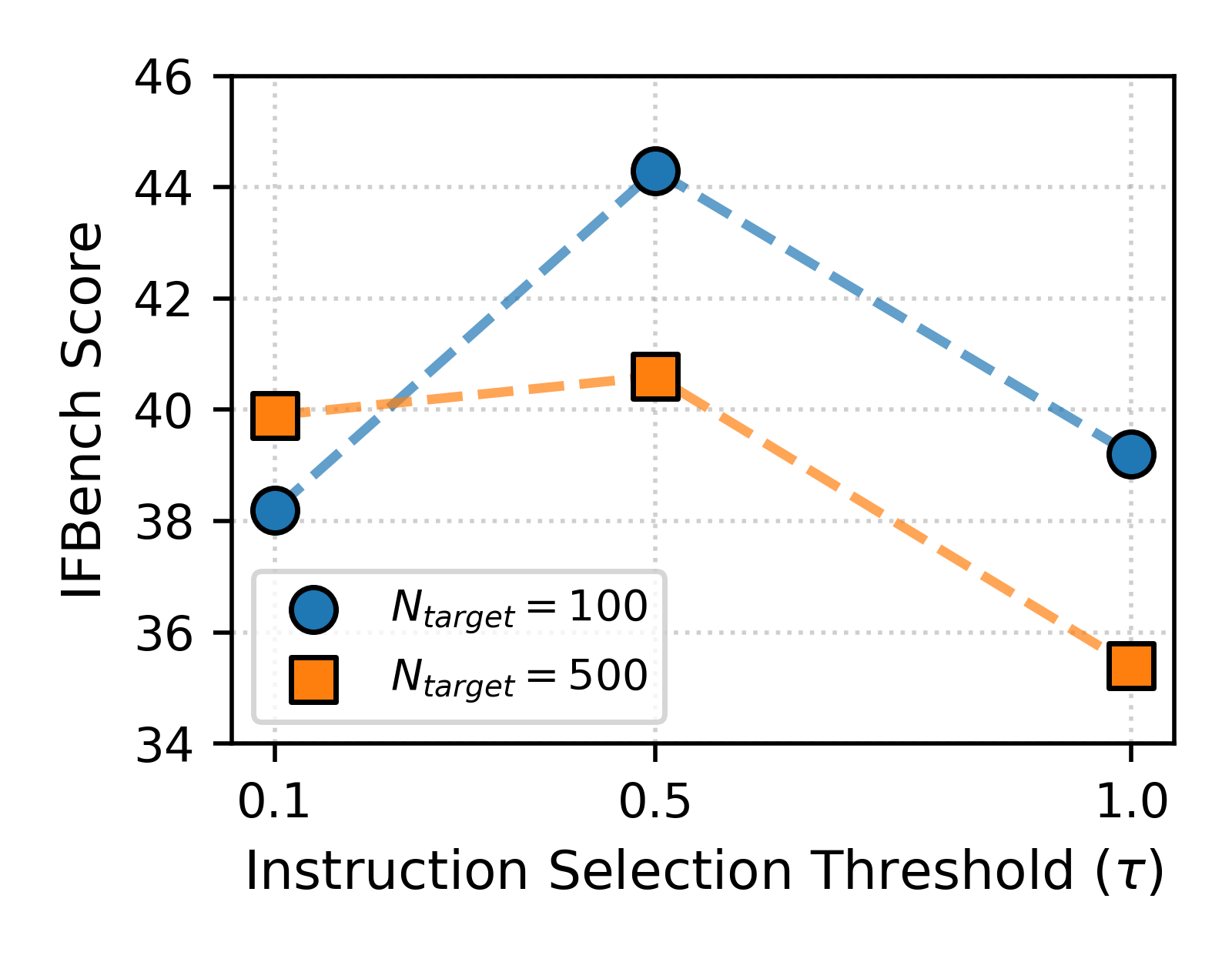}
    \label{fig:ablation_data}
  \end{subfigure}\hfill
  \begin{subfigure}[t]{0.33\textwidth}
    \centering
    \includegraphics[width=\linewidth]{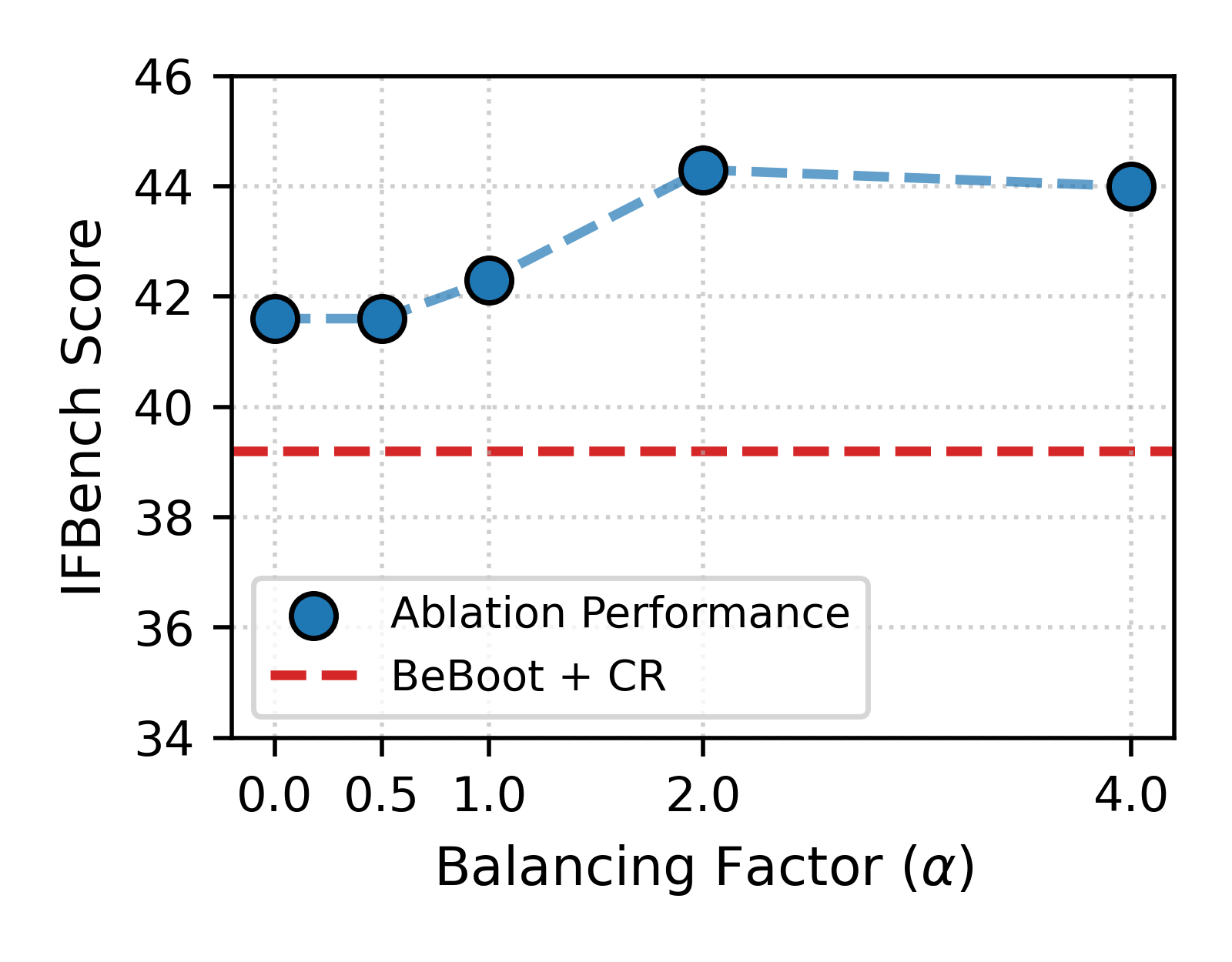}
    \label{fig:ablation_alpha}
  \end{subfigure}\hfill
  \begin{subfigure}[t]{0.33\textwidth}
    \centering
    \includegraphics[width=\linewidth]{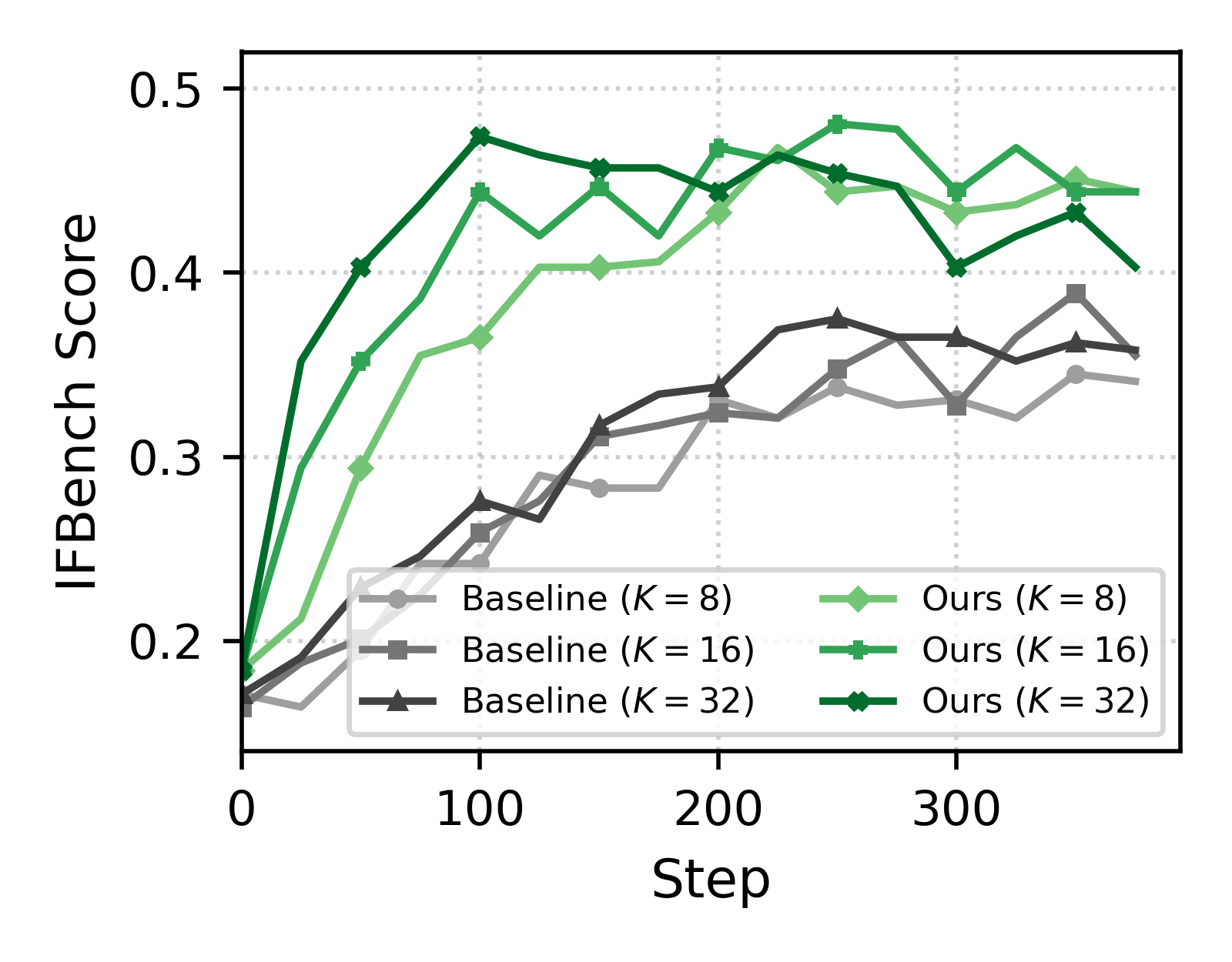}
    \label{fig:group_size}
  \end{subfigure}
  \vspace{-2em}
  \caption{Ablation results on BeBoot data (left), reward balancing factor (middle), and response group size (right).}
  \label{fig:ablations}
\end{figure*}
\paragraph{Entropy} Figure \ref{fig:entropy} visualizes the entropy of the policy models, revealing that the baseline suffers from \textit{entropy collapse}, losing the capacity to explore the sparse frontier of complex constraints. In contrast, our method maintains significantly higher entropy: the model keeps exploring diverse responses, increasing the probability of finding a "all-pass" response. This is consistent with the better prompt accuracy of our models on training and evaluation benchmarks, further demonstrating the effectiveness of BeBoot and SaR.

% proving that BeBoot effectively activates the model's ability to attempt hard instructions. The subsequent gradual decline under SaR demonstrates a stable optimization process that avoids sub-optimal convergence while mastering rare instruction combinations.

\subsection{Ablation}\label{sec:ans}
\paragraph{Impact of the Behavior BootStrapping Data}
According to our primary findings, Behavioral Bootstrapping (BeBoot) successfully activates instructions with low initial accuracy, leading to improved optimization efficiency during the reinforcement learning phase. We further investigate how the composition and scale of the bootstrapping data impact RL performance by conducting an ablation study on two critical hyperparameters: the instruction selection threshold $\tau$ and the target data count $N_{target}$. $\tau$ determines the data diversity and $N_{target}$ determines data size. The sample size $K$ for generating rollouts is fixed at 32, as we observed that if the model fails to produce a valid rollout within 32 attempts, the probability of successful generation remains exceedingly low. We evaluate $\tau$ at three levels: 0.1 (targeting hard instructions), 0.5 (targeting both hard and medium instructions), and 1.0 (targeting all the instructions). Additionally, we vary $N_{target}$ between 100 and 500 to compare the effects of small-scale versus larger-scale seed data. Figure \ref{fig:ablations} (left) illustrates the model performance under different data settings. We observe that an instruction selection threshold of $\tau=0.5$ consistently achieves superior results compared to $\tau=0.1$ or $\tau=1.0$. This indicates that \textbf{including a moderate diversity of non-easy instruction types is optimal for behavioral activation}. Furthermore, the smaller data scale ($N_{target}=100$) generally outperforms the larger scale ($N_{target}=500$). This suggests that \textbf{behavioral activation is most effective when restricted to a lightweight activation phase; excessive data likely leads to over-optimization}, which may limit the model's flexibility during the subsequent reinforcement learning stage. 

\paragraph{Analysis of the Balancing Factor}
Figure \ref{fig:ablations} (middle) illustrates the sensitivity of the model's performance to the balancing factor $\alpha$. We first observe that even when $\alpha=0$ (employing only unary rewards), the model significantly outperforms the BeBoot+CR baseline, indicating the effectiveness of the scarcity-aware reward design. As the weight of the binary reward increases, performance remains stable from $\alpha=0.0$ to $0.5$ before trending upward to a peak score at $\alpha=2.0$. This trend underscores that while small values may have little impact on training and excessively large values can lead to over-optimization of instruction combinations, properly balancing unary and binary rewards can effectively optimize the model performance.

\paragraph{Response Group Size} The scarcity-aware rewards are calculated based on the entire response group. We study the effect of the response group size $K$ by setting it to 8, 16, and 32. As shown in Figure \ref{fig:ablations} (right), our method robustly improves upon the corresponding baselines with the same group sizes. While larger group sizes lead to faster convergence, they may cause over-optimization, which we attribute to assigning too large rewards to rare instructions due to the large group size.
\section{Conclusion}
This work studies exploration bias in multi-instruction following with RL, where models prioritize easy instructions over complex ones. We introduce metrics to quantify the imbalanced ability of models and propose a two-stage framework with a lightweight behavioral bootstrapping phrase to activate difficult instructions and scarcity-aware rewards to incentivize rare instruction satisfaction during RL training. Our results show the framework enables small models to match the performance of larger off-the-shelf models.

\section*{Limitations}
While our methods significantly mitigate exploration bias in multi-instruction following, several limitations remain:
\begin{itemize}
    \item Our Scarcity-Aware Rewards design currently focuses on unary and binary rewards to optimize for individual instructions and pairwise synergies. This approach is based on the intuition that fulfilling a complex set of instructions can be effectively decomposed into mastering various instruction pairs. However, this may not explicitly address higher-order dependencies where three or more instructions interact simultaneously, potentially leaving some co-satisfaction bottlenecks unaddressed in extremely dense constraint scenarios. Theoretically, our framework can be generalized to $n$-ary scarcity rewards, which we leave for future investigation.
    \item Our framework requires identifying difficult instruction for behavior bootstrapping, which makes verifiable instruction following a natural testbed, where each instruction is associated with an explicit instruction type. For more open-ended instructions, instruction types could in principle be discovered via clustering, but reliable difficulty estimation and reward assignment would still depend on model-based verifiers. Because such verifiers are often less reliable and can introduce noise, we validate our framework only on verifiable instruction following in this work. Extending the method to broader open-ended instructions (e.g., style, politeness, and factuality) is an important direction for future research.
\end{itemize}

% \section*{Acknowledgments}

\bibliography{IF,yyq}

\appendix

% \section{Example Appendix}
% \label{sec:appendix}

\end{document}